# Kernel Based Cognitive Architecture for Autonomous Agents

Alexander Serov

Research Group "Automatic Intelligent Data Acquisition" (RG AIDA)
alexser1929@gmail.com

One of the main problems of modern cognitive architectures is an excessively schematic approach to modeling the processes of cognitive activity. It does not allow the creation of a universal architecture that would be capable of reproducing mental functions without using a predetermined set of perceptual patterns. This paper considers an evolutionary approach to creating a cognitive functionality. The basis of our approach is the use of the functional kernel which consistently generates the intellectual functions of an autonomous agent. We consider a cognitive architecture which ensures the evolution of the agent on the basis of Symbol Emergence Problem solution. Evolution of cognitive abilities of the agent is described on the basis of the theory of constructivism.

## 1. Introduction

The main perspective at the creation of robotic systems of the future is to achieve a cognitive level comparable to that of a human. Choosing right methodology is an essential element in solving this problem. Autonomous learning methods are currently underdeveloped. In fact, there is currently no foundation for self-learning from the scratch. In part, this can be explained by the lack of a methodology capable of adequately describing such a task. From the philosophical and psychological points of view, the ideas of *Artificial Subjective Reality* can serve as such a foundation. Living beings from the very moment of birth exist within the boundaries of individual reality, the construction and transformation of which they are engaged in throughout their entire life. From the mathematical and technical points of view, mentioned foundation can be presented as a kind of universal kernel. Its cognitive functionality must be capable of building a hierarchically organized representation of the world of agent embodiment.

The use of the ideas of Artificial General Intelligence (AGI) in the field of robotics would have significant prospects when setting tasks from the Universal Computational Intelligence class. We consider it fundamentally possible to create a universal kernel, which would be the basis for building cognitive autonomous systems capable of purposeful activity in a wide range of applications and a wide range of worlds of embodiment. Currently, the methodology of AGI is actively used when creating cognitive architectures [1], [2]. Following projects may be mentioned as examples. Learning Intelligent Distribution Agent [3] is based on the implementation of the Global Workspace Theory [4]. The Connectionist Learning with Adaptive Rule Induction On-line project [5] implements the integration of models developed based on the results of research in the field of psychology. The project related to the development of the Adaptive Control of Thought - Rational architecture [6] was developed on the basis of the Rational Analysis methodology.

From the point of view of the universality of application, modern projects have several significant drawbacks. These shortcomings are mainly determined by the insufficient level of the development of modern machine learning methods which may be used in autonomous systems. Reinforcement learning can serve as the basis for learning of autonomous agents in data streams. These methods make possible implementing a certain set of computational functions. But they are not able to provide the solution for the fundamental problems of the emergence of intelligence. The development in the field of Biology Inspired Cognitive Architectures (BICA) are associated with attempts to simulate intellectual functions based on modern knowledge accumulated in such fields of science as biology, neurophysiology, and psychology. However, modern BICA projects may be characterized by the excessively "mechanistic" representation of cognitive processes. This may be explained by the lack of a solution of *Symbol Emergence Problem* (SEP) [7], which was originally formulated as the *Symbol Grounding Problem* [8]. The basis for developing cognitive systems of the future is the thesis that these systems should operate not with symbols, but with meanings (Chinese Room argument formulated by Searle). The Symbol Emergence Problem solution involves uncovering the way of creating a system of primary symbols. According to the analysis given in [9], SEP can be considered only if the agent uses internal resources that do not have a semantic component.

Actual lack of SEP solution is one of the main problems inhibiting the development of AGI field. This paper is devoted to the creation of a cognitive architecture on the basis of a universal functional kernel. This kernel ensures the cognitive evolution of the autonomous agent throughout the entire period of its existence.

## 2. Subjective Reality in Terms of Semiotics

Our work is based on the ideas of Artificial Subjective Reality, which can best be illustrated with use of semiotics. Simulating mental phenomena requires the use of an adequate descriptive means. Requirements for these means include the ability to consider mental phenomena in dynamics from the very beginning of sensory data processing by the autonomous agent (early postnatal ontogenesis). Semiotics gives a universal terminology introduced to explain the phenomena of the subjective world.

The world of the agent's embodiment, described by the term *Umwelt* [10], is different from the real world. Umwelt is the self-centered world of an organism, the world as known, or modeled [11]. Umwelt is a subjective "slice" of the real world belonging to a representative of a particular biological species. The Umwelt of a living being consists of the signs. Agent receives the data about these signs from sensors. These signs agent interprets based on experience, knowledge and the context of the tasks being solved. Alive being has the set of sensors that characterize the state of the Umwelt and the set of effectors that are used to make the change in Umwelt.

Umwelt is always unique because it characterizes a certain living being. Umwelt contains two parts, allocated from the surrounding reality by the sense organs and the organs of action: the perceptual world (*Merkwelt*) and the operational world (*Werkwelt*). Merkwelt and Werkwelt interact through the functional cycle: *semoisis*. Semiosis is a "process in which something is a sign to some organism" [11]. The cognitive process is characterized by the creation of the *Innenwelt*. While Umwelt is a personified part of objective reality, Innenwelt is "the world as represented in the sign system of an organism" [11].

Cognition of the world by a living being may be represented by the interaction of Merkwelt, Werkwelt and Innenwelt. In our work, we use the ideology of constructivism. *Constructivist AI* [12] is based on the Constructivist Psychological Theory proposed by Piaget [13]. He created psychological constructs, on the basis of which ones it becomes possible to simulate cognitive functions. The most important question, which has not yet been answered in Constructivist AI projects, is the question of the emergence of the primary set of symbols. The solution of SEP is fundamental to AGI as it provides an opportunity to explore and control the dynamics of cognitive development.

## 3. Functional Kernel

We assume that the functions of the cognitive architecture (agent's control system) can be split into two levels. First of them is directly related to the Merkwelt and Werkwelt. We can call it the Umwelt interface layer. Data processing realized in this level depends on the embodiment of the agent. This level of architecture implements the functions of perception and direct control of the body. The second level of the architecture is not directly related to agent embodiment. Methods of data processing at this level are abstracted. They are intended for the implementation of higher mental functions. Functions of this level are directly related to Innenwelt. We can call it the mental function layer of the architecture.

In our work, we put forward the hypothesis of the existence of a universal cognitive kernel. According to this hypothesis, there is a certain set of algorithms built into the cognitive architecture and starting their work at the moment of the beginning of postnatal ontogenesis. The work of these algorithms should ensure the fulfillment of two main tasks: to maximize the probability of the agent's survival and to ensure the beginning of the cognitive process.

Innenwelt can be regarded as an intermediate link between Merkwelt and Werkwelt. All behavioral patterns and other models representing autonomous agent knowledge are interconnected parts of the Innenwelt. According to the constructivist approach, the Innenwelt is empty at the initial moment of postnatal ontogenesis. The agent at this moment is not able to act purposefully. Algorithms that should help an agent to survive can be viewed as the implementation of unconditioned reflexes inherent in natural agents (we can call them the survival reflexes).

The beginning of the cognitive process is provided by algorithms, the analogue of which in natural agents is the orienting reflex. We consider the cognitive evolution of an autonomous agent as a process of gradual formation of the Innenwelt. It is provided by the work of all the algorithms mentioned above. The work of the cognitive kernel at the primary stage of postnatal ontogenesis leads to the solution of SEP: extraction of the set of Umwelt features and adaptation of agent behavior to these features automatically leads to the creation of a set of primary symbols.

## 4. Cognitive Architecture

We have developed the architecture of a control system based on the use of the cognitive kernel. The architecture includes the following basic components: Perceptive, Motor, Intelligent, Emotional, and Volitional components. *Perceptual component* provides cognitive agent with data about Merkwelt. Motor component is responsible for the transferring of information between core components of architecture, and low-level control of agent's body. By the term "low level control" we mean this control is not directly connected with mental processes. Low-level control includes the following functions: automatic control of the processes responsible for the existence of agent's body; control of effectors that are used for manipulations in Werkwelt.

Perceptual and Motor components are those architectural elements that provide the existence of the agent and its interaction with reality. These two components implement the Umwelt interface layer of the agent control system. The process of cognition runs via the experience of Merkwelt perception and the experience of action in Werkwelt. It leads to creation of knowledge and its consolidation into certain patterns of behavior. Intelligent component is responsible for creating, modifying, storing and managing mental models (symbols) that describe reality. Intelligent component implements Innenwelt.

According [14], the emotions of a natural agent are integral characteristics. Each emotion is a reflection of the quality and magnitude of the need experienced by the agent. Every emotion also depends on the possibility of need satisfaction, which is assessed by the agent on the basis of individual experience. *Emotional component* is responsible for two interconnected aspects of data/information processing: agent learning, and patterns recognition. Autonomous agent learns in a self-supervised manner, and emotional component is a key element in this process. At the end of the natal ontogenesis, natural autonomous agent has an unformed Innenwelt. Emotional component participates in creating and categorizing hierarchically organized set of mental models. The process of Innenwelt creation includes programming of behavioral reactions. This programming is based on the categorization of high level models. Emotional component is responsible for the creation of categories. It is also responsible for assigning labels of categories to mental models activated during processing of sensory data. Cognition is the result of activities that have specific goals. Goal-setting, actualization of behavior patterns, control of the results of actions are carried out due to the architecture component that performs the function of general management.

*Volitional component* coordinates the work of other core components of architecture. It is responsible for allocation of energy resources. The emotional component is involved in the processes of mental activity, performing four types of functions [14]. The reinforcing function of emotions is manifested in the process of the formation of conditioned reflexes. In such situations, reinforcement does not serve to satisfy the need, but to obtain desirable or remove unwanted stimuli. The reflective-evaluative function of emotions is associated with an integral assessment of the relevance and strength of the need experienced by the agent, as well as the likelihood of its satisfaction. This function is one of the elements of the agent learning and decision-making processes. The switching function of emotions is revealed in the process of competition between the motives of the agent's actions. The decision-making process is associated with the selection of a dominant need from a certain set under the influence of emotional assessment. The compensatory (substitution) function of emotions is associated with the special position that emotions occupy in the processes of mental activity.

As an active state of specialized brain structures, emotions affect the functioning of systems that regulate behavior, sensory processing, and autonomous functions. The purpose of emotions, in this case, is to weaken the requirements for assessing incoming stimuli and to mobilize the body's resources for a possible response. Typical reaction occurs in a new or unusual environment in which Innenwelt is unable to provide the agent with sufficient information to predict and evaluate. Emotions do not carry information about the environment, but in this case they substitute the existing lack of information in order to adapt the agent to the changed conditions of existence.

We describe the dynamics of the cognitive architecture using the *Schema Mechanism* [15], which has been modified in accordance with the features of our architecture. Each basic component of the architecture creates its own set of schemas. These schemas are then modified, expanded and linked with the schemas of other components, leading to holistic behavioral schemas of the autonomous cognitive agent.

## 5. Discussion

The distinctive features of the proposed cognitive architecture are as follows. *First*, the initial set of schemas may be formulated on the basis of a set of reflexes built into the cognitive architecture. However, we do not introduce a symbolic definition for Merkwelt and Werkwelt. Perceptual symbols and action symbols are determined in the course of cognitive activity. This means that the Symbol Grounding Problem is solved during the Innenwelt creation process. *Second*, according to the constructivist approach, at the initial moment of postnatal ontogenesis, the schemas associated with the level of mental functions are absent in our approach. These schemas are the result of the cognitive evolution of the initial set of schemas that are associated with motor and emotional components. *Third*, in our method we introduce schemas that are directly related to cognitive activity. They are introduced using the orientation reflex. These schemas, being themselves subject to evolution, also participate in the creation of schemas used in the implementation of functions of the mental level.

Developmental psychology studies [16] show that the most significant changes in Innenwelt occur at the sensorimotor stage, which is the first in the postnatal ontogenesis of natural agents. This stage is characterized by the beginning of the formation of a set of symbols of perception and symbols of action. According to Piaget's views, this period of development is one of the most important in the formation of a person's mental abilities. Proposed architecture makes it possible to simulate the process of cognitive evolution, *including the stage of sensorimotor development* of autonomous agents. This problem is solved on the basis of the set of unconditioned reflexes. We consider the proposed approach as a development of Drescher's ideas. It allows one to use the results of research in the field of psychology of early child development and creation of relevant numerical models to simulate the cognitive activity of autonomous agents throughout their life cycle.

The activity of an agent can be phenomenologically characterized as purposeful from the moment when the following conditions are met. The spaces of symbols of perception and symbols of action are not empty. The agent realizes that it is capable of manipulating the action symbols. A mapping from the agent's needs space to the action symbol space has been created. Innenwelt provides forward and backward mapping between perceptual and action symbol spaces. As a result of these conditions being met, the agent becomes able to formulate the goals of its actions in terms of symbols of perception and scenarios of its behavior in terms of symbols of action.

Piaget expressed the confidence that the complication of the methods used by people in their activities and the complication of methods of representing (reflecting) reality are two fundamentally inseparable aspects of the process of cognitive development. The transition to modeling the cognition process (starting from its earliest stages) makes it possible to study the patterns of this process and use them to develop autonomous systems with a given set of characteristics. A detailed study of the dynamics of the development of mental functions, starting from the sensorimotor stage, from our point of view, is the only direction that leads to the creation of *true* autonomous AGI systems.